# Beyond Bayer: Task-Optimal Sensor Co-Design for Robust Autonomous-Driving Segmentation

Reeshad Khan[1] and John Gauch[1]

University Of Arkansas, Fayetteville AR 72701, USA

**Abstract.** Robust perception underpins autonomous driving, and most recent progress comes from scaling the *model*-larger backbones, foundation models, and cooperative multi-agent fusion. We pursue a complementary, upstream question: what should the *camera* itself measure? Using a differentiable RAW-to-task pipeline, we decompose which sensor degrees of freedom benefit dense prediction. Learning the spectral colour-filter-array (CFA) weights is the dominant lever, improving mIoU by +0.017 (KITTI-360) and +0.023 (ACDC) over a fixed camera. In contrast, point-spread-function (optics) co-design is net-negative (−0.020 mIoU on KITTI-360)-a consequence of the data-processing inequality, which also bounds the task information that any downstream model, however large or cooperative, can recover. Noise co-optimisation is marginal, and-counter to intuition-enlarging the CFA tile beyond 2×2 consistently hurts, as the filters are confined to the rank-three sRGB input. Because the intervention is at the sensor, the gains are model-agnostic; we validate robustness on ACDC's fog, night, rain, and snow, and conclude with a simple recipe: learn the 2×2 CFA weights and keep an identity PSF.

**Keywords:** Autonomous driving · Robust perception · Sensor co-design · Semantic segmentation · Computational imaging · Colour filter arrays

## 1 Introduction

Semantic segmentation is a core perception primitive for autonomous driving, and modern transformer backbones [28] reach high accuracy on urban and adverse-condition benchmarks such as KITTI-360 [16] and ACDC [23]. Yet the sensor that feeds these networks is rarely designed with them in mind. A conventional camera-an aberration-corrected lens followed by a Bayer colour-filter array tuned for perceptual colour reproduction-optimises the *image* a human would want to see, not the *decision* a network has to make. The resulting pipeline is a sequence of locally sensible but globally task-agnostic choices.

End-to-end *deep optics* reframes the camera and the network as a single differentiable system that can be optimised jointly [29]. This view has produced lenses tuned for downstream recognition [30], alternating optics/post-processing schemes [20], and manufacturability-aware designs [10], and has been studied for segmentation under optical aberrations [15]. Most of this work optimises the *optics* (the point-spread function, PSF), sometimes alongside the CFA and the noise model, and reports gains on classification, detection, or image quality. The implicit premise is that every sensor degree of freedom is worth co-optimising.

We argue that this premise is false for dense prediction, and that the sensor's degrees of freedom split into a sharp asymmetry. On one side, PSF co-design is *net-negative* for segmentation. A passive lens can only attenuate spatial frequencies, so under any noise floor it can only discard information that the network needs at every pixel; the data-processing inequality makes the identity lens information-optimal, and joint optimisation of the PSF therefore ranges from neutral to harmful rather than helpful. On the other side, the spectral CFA is a genuine lever: a $2\times2$ Bayer pattern allocates its filters for percep-tual luminance, not for separating semantic classes, and re-allocating them toward class-discriminative spectral mixtures increases the task-relevant information the sensor captures without changing the per-pixel measurement budget. This asymmetry-spatial degrees of freedom can only lose, spectral degrees of freedom can gain-reframes what is worth optimising in a co-designed sensor.

Acting on it, we build a differentiable RAW-to-task pipeline with a learnable $T\times T$ spectral CFA and use it as an analysis instrument: we fix the optics to identity and ask which CFA configuration is best for the task. The answer is unexpectedly simple. The optimum is a learned $2\times2$, Bayer-structured CFA; enlarging the tile to $3\times3$ or $4\times4$ *consistently degrades* accuracy on both datasets. The reason is informational: the input remains three-channel sRGB, so the filters re-mix existing colour information rather than acquiring new spectral bands, and once three independent directions are spanned, extra tile sites cannot add spectral information-they only coarsen spatial sampling, which dense prediction is sensitive to. We make no hyperspectral claim; methods that acquire new bands require a hyperspectral input camera [24, 25], whereas we target the commodity RGB-sensor regime.

Our contributions are:

- **An information-theoretic account of sensor co-design for dense prediction.** We show that, under any noise floor, the identity PSF maximises the task-relevant information reaching the network, so optics co-design cannot benefit segmentation, while the spectral CFA is a genuine lever (Sec. 2.3).
- **A differentiable spectral-CFA pipeline.** We instantiate a learnable $T\times T$ CFA with a Poisson–Gauss noise model and soft demosaicking, trained end-to-end through a SegFormer-B4 backbone, that lets us isolate the contribution of each sensor dimension (Sec. 2.4).
- **A controlled decomposition on KITTI-360 and ACDC.** We separate the effects of CFA learning, noise co-optimisation, optics, and CFA tile size, and find CFA learning to be the dominant gain ($+0.017/+0.023$) and optics co-design net-negative ($-0.020$ on the matched-physics KITTI comparison) (Sec. 4).
- **A simple, theory-grounded recipe, including a negative result.** For standard RGB-input segmentation the optimal sensor co-design is to learn a 2×2 CFA and keep an identity PSF; larger spectral tiles consistently hurt-a negative result with a clear information-theoretic interpretation (Sec. 5).

*Relation to foundation models and cooperative perception.* The dominant trends in driving perception-foundation models and V2X cooperative fusion-operate *downstream* of the sensor, on whatever pixels the camera delivers. Our analysis is orthogonal and prior to both: Proposition 1 shows the sensor sets an information ceiling that no backbone, however large, and no cross-agent fusion can exceed. A task-optimal sensor therefore raises the ceiling those methods operate under, and our findings compose with them rather than compete. We deliberately keep the study single-sensor to isolate the sensor question; extending the analysis to multi-camera and infrastructure-mounted (V2I) sensing is a natural next step.

## 2 Method

We study sensor–model co-design for dense semantic segmentation. Our central claim is that the three classical degrees of freedom of an imaging sensor-point-spread function (PSF/optics), colour filter array (CFA), and photon/read noise-are *not* equally useful to a dense-prediction task. We first formalise the RAW-to-task pipeline (Secs. 2.1 and 2.2), then give an information-theoretic analysis showing that PSF co-design cannot raise the task-relevant information available to a dense predictor, whereas spectral (CFA) design can (Sec. 2.3). This analysis motivates our study of a learnable spectral CFA (Sec. 2.4): we fix the PSF to identity and replace the fixed 2×2 Bayer pattern with a learnable $T \times T$ spectral CFA optimised directly for segmentation, treating the tile size $T \in \{2, 3, 4\}$ as a variable whose optimum we determine empirically (Sec. 4).

### 2.1 RAW-to-Task Pipeline Overview

Let $\mathbf{x} \in \mathrm{R}^{H \times W \times 3}$ denote the latent linear sRGB scene radiance and $\mathbf{s} \in \{1, \ldots, K\}^{H \times W}$ the semantic label map. A differentiable sensor (Fig. 1) maps $\mathbf{x}$ to a RAW mea-surement $\mathbf{y}$, which is soft-demosaicked to a feature image $\tilde{\mathbf{x}} = g_\phi(\mathbf{y})$ and passed
to a segmentation backbone $f_\vartheta$ producing $\hat{\mathbf{s}} = f_\vartheta(\tilde{\mathbf{x}})$. The full forward model is

$$\mathbf{y} = \mathrm{M}(\mathrm{N}(e \cdot (p * \mathbf{x}))), \quad \hat{\mathbf{s}} = f_\vartheta(g_\phi(\mathbf{y})), \tag{1}$$

where $p$ is the PSF, $e$ an exposure/gain scalar, $\mathrm{N}$ the sensor noise process, and $\mathrm{M}$ the CFA mosaicking operator. All operators in Eq. (1) are differentiable, so the sensor parameters $(p, \mathrm{M})$ and the network parameters $(\phi, \vartheta)$ can in principle be optimised jointly end-to-end. The remainder of this section asks which of these sensor parameters are worth optimising at all.

### 2.2 Physical Sensor Model

*Optics.* The PSF $p$ is a normalised, non-negative kernel ($p(\mathbf{u}) \geq 0, \int p \, \mathrm{d}\mathbf{u} = 1$) modelling spatially-invariant incoherent optical blur; its action $p * \mathbf{x}$ is parameterised through a differentiable optical model in the spirit of DeepLens [29]. The

Task-Optimal Spectral Mosaic Co-Design Pipeline

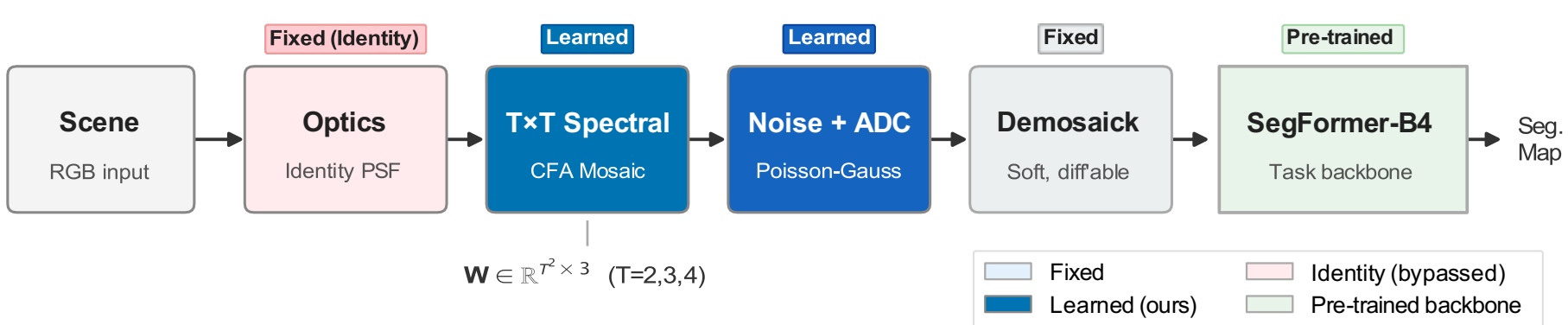

**Fig. 1:** The differentiable RAW-to-task pipeline. A scene passes through an identity PSF (optics are not co-designed), a learnable $T \times T$ spectral CFA mosaic, a Poisson–Gauss noise and ADC model, and a soft differentiable demosaicker, before a SegFormer-B4 backbone predicts the semantic map. Gradients from the segmentation loss reach the CFA and noise parameters; the optics stage is fixed to identity for the reasons given in Sec. 2.3.

non-negativity and normalisation are not modelling conveniences but physical constraints: an incoherent intensity PSF is an energy distribution.

*Colour filter array.* The CFA samples one spectral projection per pixel. A tile of size $T$ defines $T^2$ learnable spectral responses $W = \{\mathbf{w}_i\}_{i=1}^{T^2}$, $\mathbf{w}_i \in \mathbb{R}^3$, tiled periodically over the sensor plane. At pixel $\mathbf{u}$ with tile index $i(\mathbf{u})$, the mosaicked value is the scalar projection $y(\mathbf{u}) = \mathbf{w}_{i(\mathbf{u})}^{\top} \tilde{x}(\mathbf{u})$, where $\tilde{x}$ is the (optionally blurred, noisy) radiance. The standard Bayer array is the special case $T = 2$ with three distinct filters $\{R, G, B\}$ and the green response repeated, tuned for perceptual colour reconstruction rather than for class discriminability.

*Noise.* $\mathcal{N}$ models signal-dependent photon shot noise and signal-independent read noise, followed by quantisation. The key property we use is that $\mathcal{N}$ imposes a strictly positive noise floor: the measurement has finite SNR at every spatial frequency.

### 2.3 Information-Theoretic Analysis of Sensor Dimensions

We treat the sensor as a noisy channel between the scene and the network and ask how each degree of freedom affects the task-relevant information that *any* downstream network can recover. Throughout, $I(\cdot\,;\cdot)$ denotes mutual information.

*The sensor sets a ceiling.* Because the prediction depends on the scene only through the measurement, $\hat{\mathbf{s}} = f_\theta(g_\phi(\mathbf{y}))$ is conditionally independent of $\mathbf{s}$ given $\mathbf{y}$. The data-processing inequality then gives the following, regardless of the choice of $f_\theta$ or $g_\phi$.

**Proposition 1 (Sensor information ceiling).** *For the dependence structure* $\mathbf{s} \leftarrow \mathbf{x} \rightarrow \mathbf{y} \rightarrow \hat{\mathbf{s}}$ *with* $\hat{\mathbf{s}}$ *a (possibly stochastic) function of* $\mathbf{y}$*,*

$$I(\hat{\mathbf{s}}\,;\mathbf{s}) \;\leq\; I(\mathbf{y}\,;\mathbf{s}) \;\leq\; I(\mathbf{y}\,;\mathbf{x}). \tag{2}$$

*No demosaicking, backbone, or post-processing can recover more task information than the sensor measurement contains.*

Proposition 1 reframes the design problem: rather than asking what a network can do, we ask which sensor maximises the ceiling $I(\mathbf{y};\mathbf{x})$ (and, for the task, $I(\mathbf{y};\mathbf{s})$).

**PSF: the data-processing limit forces identity.** A common intuition is that a learned PSF could "encode" useful structure for the network. The following shows this cannot help a measurement-limited dense task: under any fixed noise floor, the identity PSF is the unique maximiser of $I(\mathbf{y};\mathbf{x})$ among physically realisable optics. We analyse a locally-stationary Gaussian model of the radiance as a tractable proxy; the non-negativity argument it relies on is exact and model-free.

**Proposition 2 (Identity is the information-optimal passive PSF).** *Let* $\mathbf{b} = p * \mathbf{x}$ *with* $p$ *non-negative and normalised* ($\int p = 1$)*, and let* $\mathbf{y} = \mathbf{b} + \mathbf{n}$*, where* $\mathbf{x}$ *and* $\mathbf{n}$ *are independent zero-mean stationary Gaussian fields with power spectral densities* $S_{\mathbf{x}}(\omega) \geq 0$ *and* $S_{\mathbf{n}}(\omega) > 0$*. Then for every spatial frequency* $\omega$ *the optical transfer function obeys* $|\hat{p}(\omega)| \leq 1$*, with equality for all* $\omega$ *if and only if* $p = \delta$*; and the mutual-information rate*

$$I(\mathbf{y};\mathbf{x}) \;=\; \frac{1}{2}\int \log\left(1 + \frac{|\hat{p}(\omega)|^2\, S_{\mathbf{x}}(\omega)}{S_{\mathbf{n}}(\omega)}\right) d\omega \tag{3}$$

*is maximised, uniquely wherever* $S_{\mathbf{x}}(\omega) > 0$*, by* $p = \delta$ *(the identity).*

*Proof.* For non-negative $p$ with $\int p = 1$, $|\hat{p}(\omega)| = \left|\int p(\mathbf{u})\, e^{-i\boldsymbol{\omega}\cdot\mathbf{u}}\, d\mathbf{u}\right| \leq \int p(\mathbf{u})\, d\mathbf{u} = 1$, since $|e^{-i\boldsymbol{\omega}\cdot\mathbf{u}}| = 1$ and $p \geq 0$; equality at a given $\omega \neq \mathbf{0}$ requires the phase $e^{-i\boldsymbol{\omega}\cdot\mathbf{u}}$ to be constant on $\mathrm{supp}(p)$, and equality for *all* $\omega$ forces $\mathrm{supp}(p) = \{\mathbf{0}\}$, *i.e.* $p = \delta$. The integrand in Eq. (3) is the Gaussian-channel information rate, strictly increasing in $|\hat{p}(\omega)|^2$ wherever $S_{\mathbf{x}}(\omega) > 0$; substituting the maximal value $|\hat{p}|^2 \equiv 1$ maximises it. □

**Corollary 1 (PSF co-design cannot help dense prediction).** *Combining Propositions 1 and 2, for any non-degenerate PSF the attainable task information* $I(\hat{\mathbf{s}};\mathbf{s})$ *is bounded above by that of the identity-PSF sensor. Hence optimising the PSF can at best match (at* $p = \delta$*) and otherwise reduces the information available for dense prediction.*

The mechanism is explicitly the noise floor: a passive PSF can only attenuate spatial-frequency content ($|\hat{p}| \leq 1$), and any attenuated band is pushed toward

$S_n$, irreversibly losing information that deconvolution cannot restore (Proposition 1). Dense prediction depends on precisely the high-frequency, per-pixel content that such attenuation destroys. This also clarifies an apparent contradiction with task-driven lens design [30], which improves *classification*: a global label requires only a low-dimensional sufficient statistic $T(\mathbf{x})$, so a PSF can discard local spatial detail while preserving $I\big(T(\mathbf{x});\mathbf{y}\big)$ and even act as a useful optical prior. The same operation is harmful when the label is itself dense.

**Spectral CFA: diversity buys discriminability.** The spectral degree of freedom behaves oppositely. Within a tile, the sensor collects $T^2$ scalar projections $\{\mathbf{w}_i^\top \tilde{x}\}$ of the local spectral signal under a one-scalar-per-pixel budget. Treating the local radiance as approximately constant over a tile, the $T^2$ projections preserve the task-relevant spectral information best when the directions $\{\mathbf{w}_i\}$ span the discriminative subspace and are mutually non-redundant. Bayer's filters are chosen for perceptual luminance, not for separating semantic classes; reallocating them toward class-discriminative directions increases $I(\mathbf{y};\mathbf{s})$ without changing the per-pixel budget. This is the lever we optimise.

*What this does* not *claim.* The input radiance $\mathbf{x}$ is three-channel sRGB, so every learned response $\mathbf{w}_i \in \mathbb{R}^3$ lives in a rank-3 space: at most three of the $T^2$ filters can be linearly independent. Our gains therefore come from *task-aligned* spectral mixing and *denser spatial sampling* of those mixtures, *not* from acquiring new spectral bands. We make no hyperspectral claim; true spectral super-resolution would require a hyperspectral input sensor, as in [24, 25].

### 2.4 Task-Optimal Spectral Mosaic Design

Guided by Corollary 1 and Sec. 2.3, we fix the optics to identity ($p = \delta$, no PSF co-design) and optimise only the spectral mosaic and the network.

**$T \times T$ learnable CFA.** The mosaic is parameterised by an unconstrained matrix $\Omega \in \mathbb{R}^{T^2\times 3}$ whose rows are mapped to non-negative spectral responses that respect filter physics (each response a convex mixture of the input channels) by a row-wise softmax:

$$\mathbf{w}_i = \mathrm{softmax}(\Omega_i), \qquad i = 1, \ldots, T^2. \tag{4}$$

The mosaic operator $\mathsf{M}$ applies $\mathbf{w}_{i(\mathbf{u})}$ at each pixel as in Sec. 2.2. Setting $T$=2 with fixed RGGB weights recovers a standard Bayer camera; comparing it against a learnable $T$=2 CFA isolates the effect of CFA *learning*, while sweeping $T \in \{2, 3, 4\}$ isolates the effect of tile *size*. We make no a priori assumption that larger $T$ is better; Sec. 4 determines the optimum.

**Spectral diversity regularisation.** To prevent the $T^2$ filters within a tile from collapsing onto one another, we add a penalty on the pairwise similarity of the (unit-normalised) responses $\bar{\mathbf{w}}_i = \mathbf{w}_i / \|\mathbf{w}_i\|$:

$$\mathcal{L}_{\text{div}} = \frac{1}{T^2(T^2-1)} \sum_{i=j} \left( \bar{\mathbf{w}}_i^{\top} \bar{\mathbf{w}}_j \right)^2. \tag{5}$$

We stress that this regulariser cannot manufacture spectral information beyond what the rank-three sRGB input provides: at most three of the $\mathbf{w}_i$ can be linearly independent (Sec. 2.3), so for $T>2$ the additional sites are necessarily more redundant regardless of $\mathcal{L}_{\text{div}}$. This is the formal seed of the negative tile-size result in Sec. 4.4.

**Differentiable demosaicking.** The single-channel mosaic $\mathbf{y}$ is reconstructed to a dense multi-channel feature image $\tilde{\mathbf{x}} = g_{\phi}(\mathbf{y})$ by a soft, fully-differentiable demosaicker, so gradients from the segmentation loss flow back to the CFA parameters $\Omega$.

**Training objective.** We train sensor and backbone jointly by minimising the segmentation loss with the diversity regulariser:

$$\min_{\Omega,\, \phi,\, \vartheta} \mathbb{E}_{(\mathbf{x},\mathbf{s})} \left[ \mathcal{L}_{\text{seg}}\left( f_{\vartheta}(g_{\phi}(\mathbf{y})),\, \mathbf{s} \right) \right] + \lambda\, \mathcal{L}_{\text{div}}, \tag{6}$$

with $\mathbf{y}$ given by Eq. (1) at $p = \delta$, $\mathcal{L}_{\text{seg}}$ the standard per-pixel cross-entropy, and $\lambda$ the regularisation weight. Crucially, the optics term is absent: by Corollary 1 its optimum is the identity we have already imposed, and-as we show empirically in Sec. 4-retaining it as a free parameter is neutral-to-harmful rather than helpful.

## 3 Related Work

**Computational imaging and deep optics.** End-to-end "deep optics" treats the camera and the downstream network as a single differentiable system, optimising the optical element jointly with the reconstruction or task network. Sitzmann *et al*. [26] jointly optimised a diffractive lens and an image-processing network for achromatic extended depth of field and super-resolution, and the paradigm has since been applied to monocular depth and 3D detection [4], single-shot HDR [18], hyperspectral-depth imaging [1], and compound refractive lens search for detection [9, 29]. Most of this line optimises the PSF (lens surface or phase mask), and recent work refines the optimisation itself-alternating optics and post-processing [20] or adding manufacturability and tolerance regularisation [10]. Our analysis is orthogonal: rather than improving PSF optimisation, we show that for dense prediction the PSF degree of freedom is information-theoretically unhelpful, and we redirect the co-design effort to the spectral filters.

**Learned colour filter arrays and RAW-to-task ISPs.** A parallel thread learns the colour filter array and demosaicker. Chakrabarti [3] learned sensor multiplexing patterns by back-propagation; Henz *et al*. [14] jointly designed CFA patterns and demosaicking; Gharbi *et al*. [13] learned joint demosaicking and denoising. These optimise reconstruction quality (PSNR). A related line questions whether the human-oriented ISP is the right front end for recognition at all: Buckler *et al*. [2] reconfigure the pipeline for vision, Diamond *et al*. [11] train end-to-end through a differentiable ISP on RAW measurements, and Chen *et al*. [5] learn to process RAW in low light. We share this RAW-to-task philosophy but ask a sharper question-which sensor dimension is worth optimising for *dense* prediction-and report a negative result on enlarging the CFA beyond the Bayer tile.

**Sensor design for segmentation and hyperspectral driving.** Closest to our setting, Jiang *et al*. [15] study segmentation under, and co-design against, optical aberrations, while task-driven lens design [30] improves *classification* with engineered optics-a result we reconcile with ours through the dense-versus-global distinction (Sec. 5.1). On the spectral side, learned quantum-efficiency filters [24] and hyperspectral ADAS sensing [25] pursue genuine spectral super-resolution, but require a hyperspectral input camera; we instead operate within the rank-three sRGB gamut of a commodity sensor, which is precisely what bounds the benefit of larger spectral tiles.

**Semantic segmentation and robustness.** Segmentation architectures have evolved from fully-convolutional networks [17] through context aggregation [6,32] to transformer backbones [7, 27, 28]; we adopt SegFormer-B4 [28] as a representative strong backbone. Urban driving benchmarks [8, 12, 16, 19, 31] and adverse-condition datasets [21–23] have driven progress on robust perception. These works improve the *model* for a fixed sensor; we instead ask what the *sensor* should measure, evaluating on KITTI-360 [16] and ACDC [23], which span normal and adverse conditions.

## 4 Experiments

### 4.1 Datasets and Setup

We evaluate on two driving-segmentation benchmarks: KITTI-360 [16] (urban scenes, normal conditions) and ACDC [23] (adverse conditions: fog, night, rain, snow). All models use a SegFormer-B4 backbone [28], trained for 40 epochs on KITTI-360 and 60 on ACDC. The differentiable sensor (Fig. 1) comprises an identity or learnable PSF, a fixed or learnable CFA, a Poisson–Gauss noise and ADC stage, and a soft demosaicker; we report mean intersection-over-union (mIoU). We compare an *RGB oracle* (clean sRGB fed directly to the backbone, an upper bound with no sensor pipeline), a *fixed-camera* baseline (fixed Bayer, identity PSF), full *co-design* (PSF+CFA+noise), and three identity-PSF spectral configurations: *no optics* (learnable CFA+noise), *CFA only* (no noise co-optimisation), and larger tiles (*Tile-3*, *Tile-4* ).

**Table 1:** Sensor co-design results on KITTI-360 and ACDC (SegFormer-B4, mIoU). PSF/CFA/Noise: ✓=learnable, ×=fixed/identity. The RGB oracle is an upper bound with no sensor pipeline. The best *sensor* configuration per dataset is in bold; ⋆ marks our primary method, the learnable 2×2 CFA (the no-noise variant is marginally best on ACDC, ±0.003). ACDC baselines (fixed camera, co-design, oracle) use the earlier sensor physics; cross-physics comparisons are directional (see text).

| Method | PSF | CFA | Noise | Tile | KITTI-360 | ACDC |
|---|---|---|---|---|---|---|
| RGB (no sensor pipeline) | — | — | — | — | 0.6396 | 0.7374 |
| Fixed camera | × | × | × | 2 | 0.6043 | 0.6654 |
| Co-design (PSF+CFA+noise) | ✓ | ✓ | ✓ | 2 | 0.6031 | 0.6906 |
| No optics **(Ours⋆)** | × | ✓ | ✓ | 2 | **0.6229** | 0.6853 |
| CFA only (no noise co-opt.) | × | ✓ | × | 2 | 0.6214 | **0.6882** |
| Tile-3 ablation | × | ✓ | ✓ | 3 | 0.6165 | 0.6800 |
| Tile-4 ablation | × | ✓ | ✓ | 4 | 0.6133 | 0.6740 |

*Physics caveat (ACDC).* On ACDC, the fixed-camera, co-design, and RGB-oracle runs were trained under an earlier sensor-physics implementation, whereas the spectral configurations use a corrected model (pre-CFA Poisson–Gauss noise and corrected regularisation). We therefore treat any ACDC comparison that crosses this boundary-most notably the co-design (PSF) contrast-as *directional only*, and we base all quantitative PSF conclusions on KITTI-360, where every configuration shares identical physics.

### 4.2 Main Results

Table 1 and Fig. 2 report the full comparison. Learning the CFA is clearly beneficial: the learnable 2×2 CFA improves mIoU over the fixed camera by +0.019 on KITTI-360 (0.6043 → 0.6229) and +0.020 on ACDC (0.6654 → 0.6853), recovering a substantial fraction of the gap to the RGB oracle (0.6396 / 0.7374). The no-noise variant is comparable (0.6214 / 0.6882); the best learnable configuration is noise-co-optimised on KITTI-360 and noise-free on ACDC, a marginal ±0.003 difference we return to below.

In contrast, optics co-design does not help. On KITTI-360, where the comparison is matched-physics, full co-design reaches only 0.6031-below both the learnable-CFA 0.6229 and even the fixed-camera 0.6043-so introducing a learnable PSF *removes* 0.020 mIoU relative to the identity-PSF learnable CFA. This is exactly the behaviour predicted by Proposition 2 and Corollary 1: a passive PSF can only attenuate spatial frequencies, and under a fixed noise floor that strictly lowers the information available for a per-pixel task. (On ACDC, co-design scores 0.6906, but under the earlier physics noted above; we do not use it to support the PSF claim.)

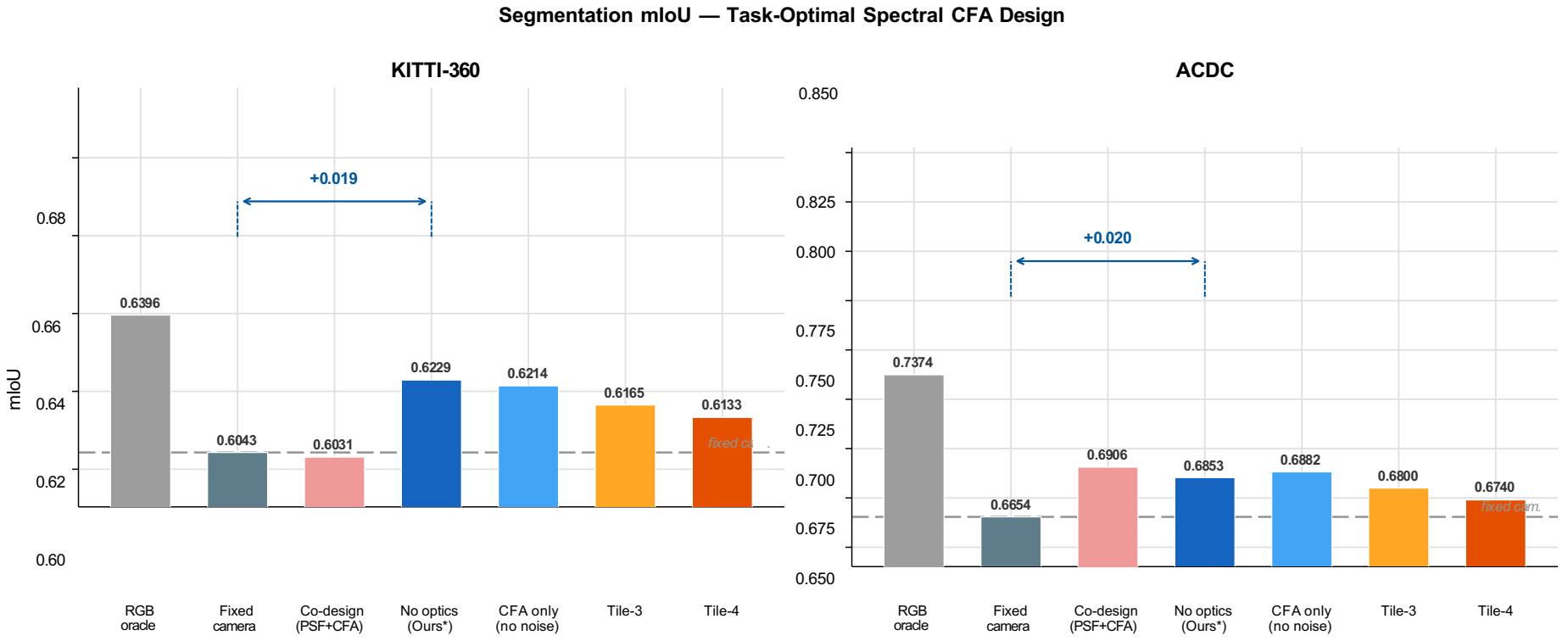

**Fig. 2:** Segmentation mIoU for all configurations on KITTI-360 (*left*) and ACDC (*right*). The learnable 2×2 CFA improves over the fixed camera (dashed line) by +0.019/+0.020. Full co-design (PSF+CFA) does not improve over the identity-PSF learnable CFA, and larger tiles (Tile-3, Tile-4) trail the 2×2 design. ACDC co-design uses the earlier physics (Sec. 4.1)

## 4.3 Sensor-Dimension Decomposition

Figure 3 decomposes the change in mIoU relative to the fixed camera into three sensor dimensions. CFA learning is the dominant, consistently positive lever (+0.017 on KITTI-360, +0.023 on ACDC). Noise co-optimisation is marginal and dataset-dependent, helping slightly on KITTI-360 (+0.002) and slightly hurting on ACDC ($-0.003$); we therefore avoid any general claim that noise co-design helps. Optics is neutral-to-harmful: $-0.020$ on the matched-physics KITTI-360 comparison, with the ACDC point (+0.005) flagged as cross-physics and directional only. The ordering-spectral $\gg$ noise > optics, with optics negative-is the empirical core of the paper.

## 4.4 CFA Tile-Size Analysis

A natural hypothesis is that more spectral filter sites should capture more task-relevant information. The data falsify it. Figure 4 shows mIoU decreasing *monotonically* with tile size on both datasets: from 0.6229 ($2\times2$) to 0.6165 ($3\times3$) to 0.6133 ($4\times4$) on KITTI-360, and from 0.6853 to 0.6800 to 0.6740 on ACDC-penalties of $-0.006/-0.010$ and $-0.005/-0.011$ respectively. The $2\times2$ tile is optimal in every case. We interpret this in Sec. 5.2; it is the empirical realisation of the rank-three bound discussed in Secs. 2.3 and 2.4.

## 4.5 Learned CFA Patterns

Figures 5 and 6 visualise the learned filters. The learnable $2\times2$ CFA converges to a near-Bayer RGGB pattern-its filters approach the primaries (1, 0, 0), (0, 1, 0), (0, 1, 0), (0, 0, 1), with a diversity score of 0.831 against Bayer's 0.833-indicating that, within the sRGB colour space, the Bayer allocation of *which* primaries

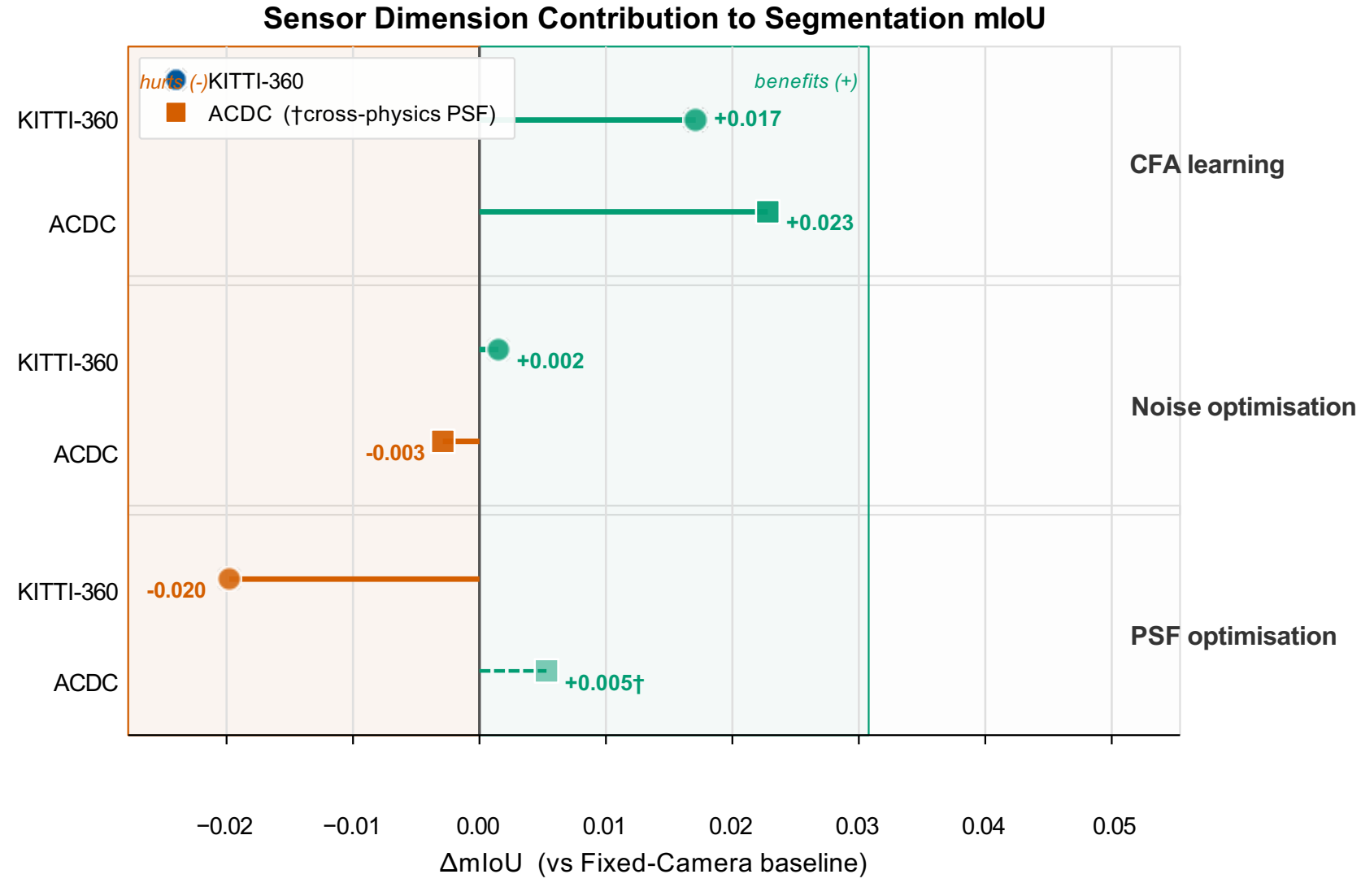

**Fig. 3:** Per-dimension contribution to mIoU relative to the fixed-camera baseline. CFA learning is strongly positive on both datasets; noise is marginal; optics is negative on KITTI-360. The ACDC optics point (†) is a cross-physics comparison and is directional only to sample is already close to optimal; the benefit of the learnable CFA comes from task-tuning the exact mixtures and training the demosaicker and backbone jointly through the sensor. The 3×3 and 4×4 designs instead discover genuinely novel broadband filters (*e.g.*, ≈ 49%R+51%G or ≈ 36%G+64%B in Fig. 6), yet their diversity is *lower* (0.554 at $T$=3, 0.465 at $T$=4): the extra sites cannot realise new independent spectral directions and are forced into redundant mixtures, exactly as the rank-three argument predicts. Figure 7 shows the corresponding spatial tilings.

## 4.6 Qualitative Results

Figure 8 compares the fixed camera and the learnable 2×2 CFA against ground truth on representative ACDC (adverse) and KITTI-360 (normal) scenes. The learnable CFA more often recovers rare and thin classes that the fixed camera misses-for instance a motorcycle that the fixed camera labels as fence, and a rider that it labels as sidewalk. The larger-tile predictions (provided in the supplementary material, with all eight scenes per dataset and the Tile-3/Tile-4 columns) show similar but less consistent recovery, mirroring the quantitative ranking.

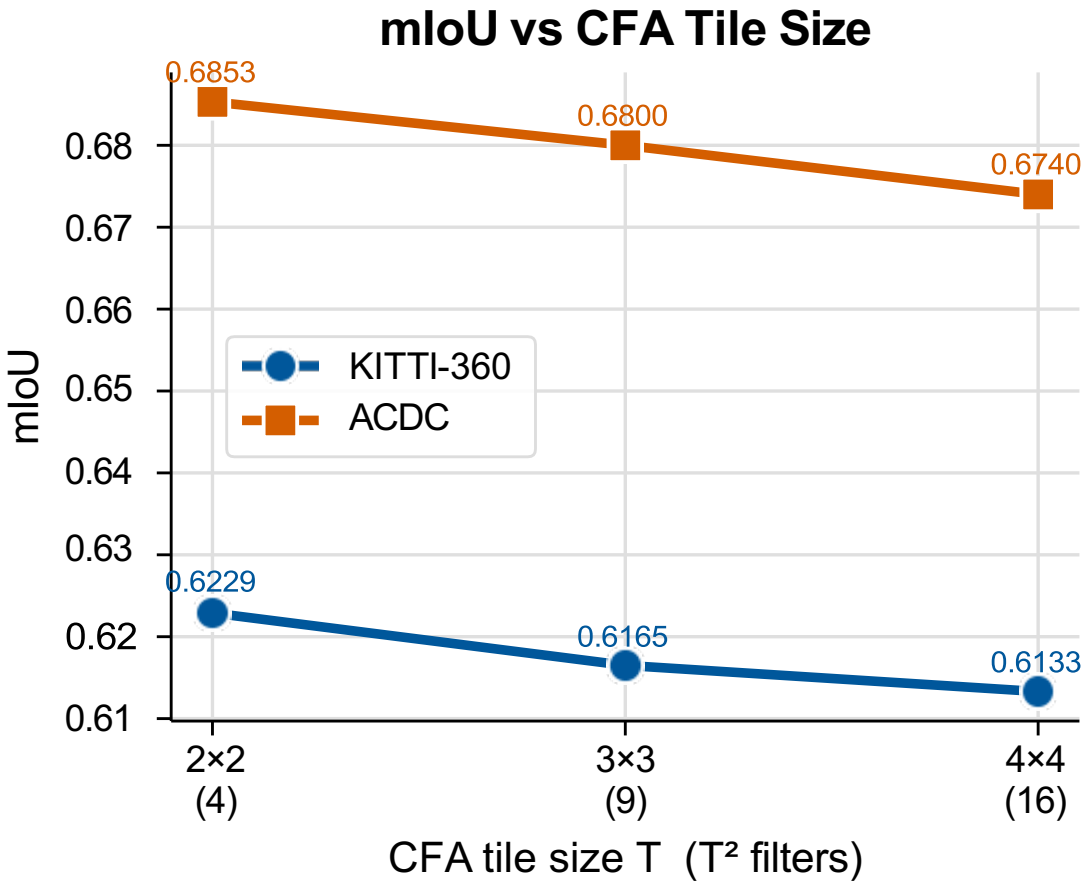

**Fig. 4:** mIoU versus CFA tile size T (with $T^2$ filter sites). Both datasets show a consistent downward trend: enlarging the tile beyond 2×2 hurts segmentation

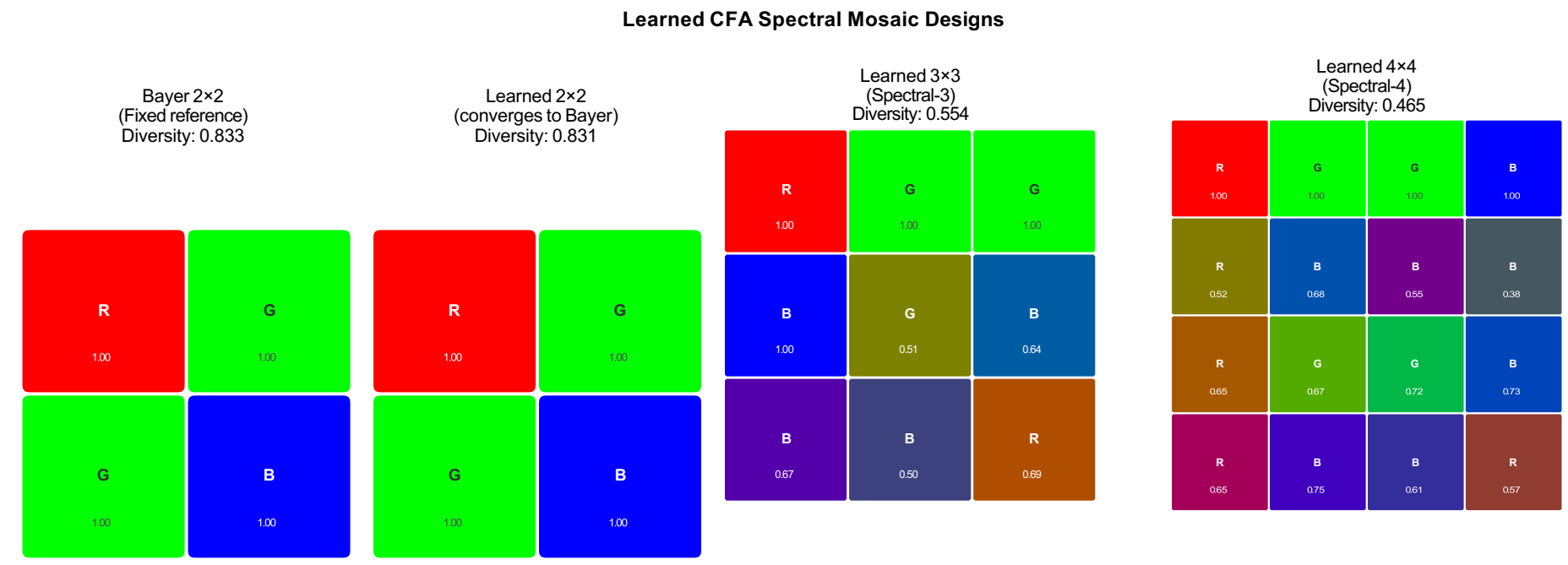

**Fig. 5:** Learned CFA tiles. The learnable 2×2 design (second) recovers a near-Bayer RGGB pattern; the 3×3 and 4×4 designs discover broadband mixed filters but with *lower* spectral diversity, reflecting the rank-three sRGB constraint

# 5 Discussion

## 5.1 When Does Optics Co-Design Help?

Our negative result for PSF co-design is specific to dense prediction and does not contradict the optics-co-design literature. By Corollary 1, the identity PSF is information-optimal whenever the task depends on per-pixel detail. A global task such as classification depends only on a low-dimensional sufficient statistic of the scene, so a structured PSF can discard spatial detail while preserving the task-relevant information-and may even serve as a useful optical prior. This is why task-driven lens design improves classification [30] while providing no benefit, and a measurable cost, for segmentation. The lesson is not that optics co-design is useless, but that its value is task-dependent and vanishes for spatially dense objectives.

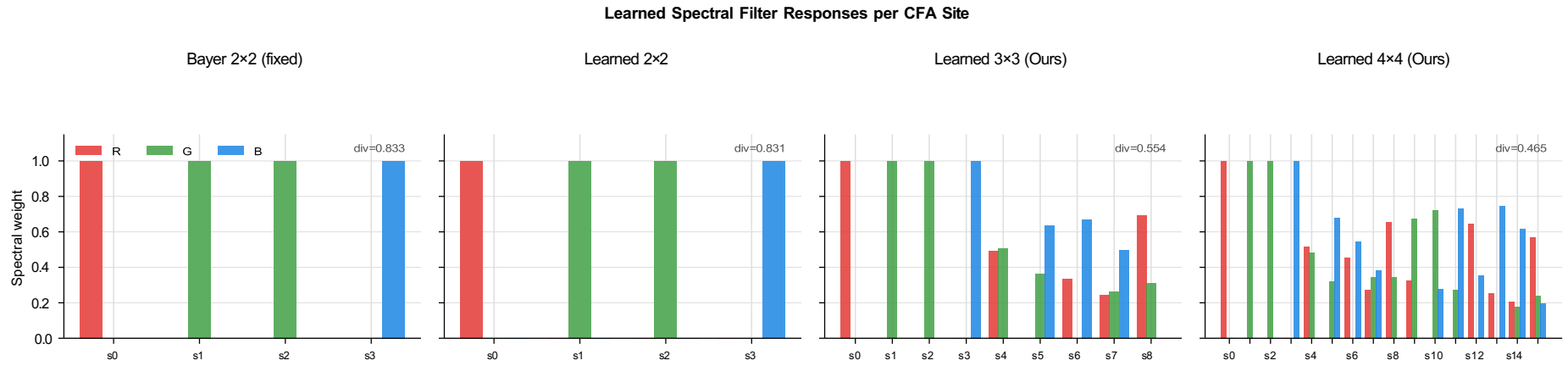

**Fig. 6:** Per-site R/G/B spectral weights for each tile size. At T=2 the sites are near-pure primaries; larger tiles use increasingly mixed, lower-magnitude responses

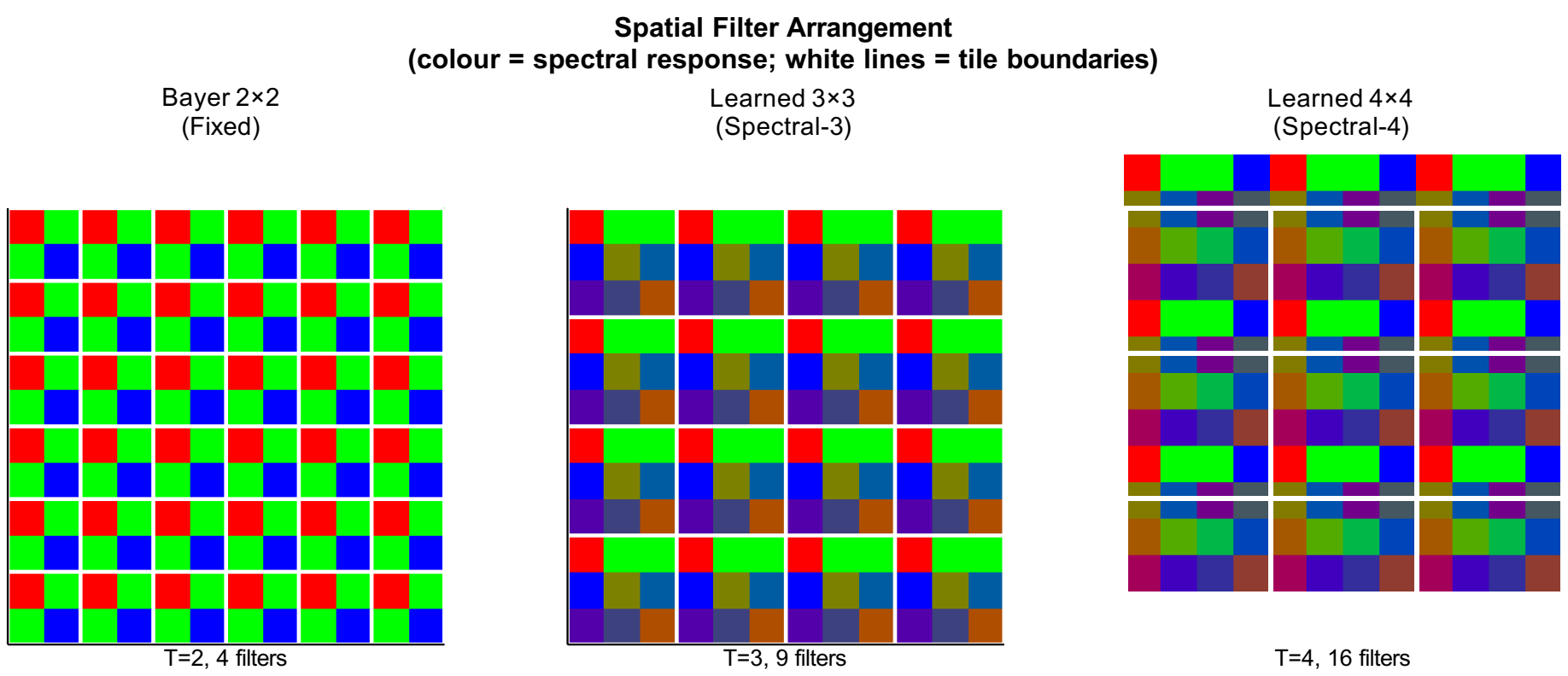

**Fig. 7:** Spatial arrangement of the learned filters across an image patch (colour = spectral response; white lines = tile boundaries) for $T \in \{2, 3, 4\}$

## 5.2 Why Does the 2×2 Tile Win?

The monotonic tile-size penalty follows from two facts. First, the input is three-channel sRGB, so the CFA can realise at most three linearly independent spectral responses (Sec. 2.3); a 2×2 tile, with four sites, already spans this space (with one repeated direction, as Bayer does). Second, every additional site spends a pixel of the periodic mosaic, so growing the tile coarsens the spatial sampling of each filter-and dense prediction is acutely sensitive to spatial resolution. Beyond $T$=2 there is thus nothing spectral to gain and spatial density to lose, which is precisely the monotonic decline we observe. The falling diversity scores ($0.83 \rightarrow 0.55 \rightarrow 0.47$) are the empirical signature of filters forced into redundancy by the rank-three ceiling.

## 5.3 Limitations and Future Work

Our filters are confined to the sRGB colour space: they re-mix existing channels and do not acquire new spectral bands, so we make no hyperspectral claim; realising genuine spectral gains would require a hyperspectral input sensor [24, 25]. On ACDC, a clean optics comparison awaits re-running the fixed-camera and co-design baselines under the corrected physics; until then the ACDC optics number is directional. Finally, our conclusions are established at SegFormer-B4

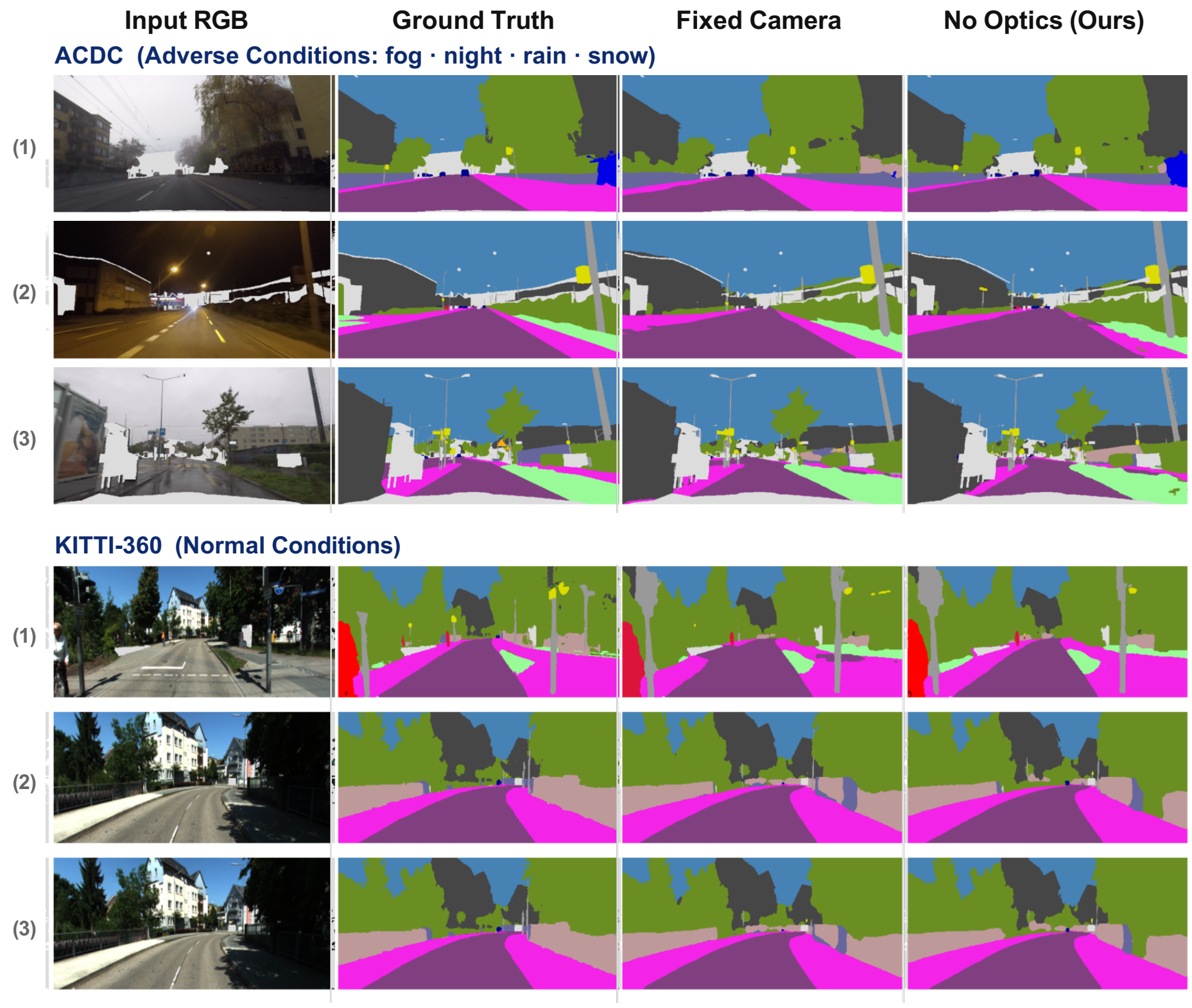

**Fig. 8:** Qualitative segmentation on ACDC (adverse) and KITTI-360 (normal). Columns: input, ground truth, fixed camera, learnable 2×2 CFA (ours). The learnable CFA recovers rare classes (*e.g.*, rider, motorcycle) that the fixed camera misclassifies

scale on two driving datasets; we explicitly do not claim that PSF co-design helps at smaller scales, and broader validation across backbones and tasks is left to future work.

## 6 Conclusion

We asked which degrees of freedom of an imaging sensor actually benefit dense semantic segmentation, and answered it with theory and a controlled decomposition. Of the three classical sensor dimensions, only re-tuning a Bayer-sized colour filter array for the task helps; optics co-design is net-negative, by the data-processing inequality, and enlarging the spectral mosaic beyond $2\times2$ consistently hurts, by the rank-three limit of an sRGB input. For standard RGB-input segmentation, the optimal sensor co-design is thus strikingly simple: learn the $2\times2$ CFA spectral weights and keep an identity PSF. Because this intervention is at the sensor, the result is model-agnostic and *upstream* of the foundation-model and cooperative-perception methods that dominate driving perception: it raises the information ceiling (Proposition 1) under which any such method must operate, and multi-camera and infrastructure-mounted (V2I) sensing are natural extensions.

## References

1. Baek, S.H., Ikoma, H., Jeon, D.S., Li, Y., Heidrich, W., Wetzstein, G., Kim, M.H.: Single-shot hyperspectral-depth imaging with learned diffractive optics. In: ICCV. pp. 2651–2660 (2021) 7
2. Buckler, M., Jayasuriya, S., Sampson, A.: Reconfiguring the imaging pipeline for computer vision. In: ICCV. pp. 975–984 (2017) 8
3. Chakrabarti, A.: Learning sensor multiplexing design through back-propagation. In: NeurIPS (2016) 8
4. Chang, J., Wetzstein, G.: Deep optics for monocular depth estimation and 3D object detection. In: ICCV. pp. 10193–10202 (2019) 7
5. Chen, C., Chen, Q., Xu, J., Koltun, V.: Learning to see in the dark. In: CVPR. pp. 3291–3300 (2018) 8
6. Chen, L.C., Zhu, Y., Papandreou, G., Schroff, F., Adam, H.: Encoder-decoder with atrous separable convolution for semantic image segmentation. In: ECCV (2018) 8
7. Cheng, B., Misra, I., Schwing, A.G., Kirillov, A., Girdhar, R.: Masked-attention mask transformer for universal image segmentation. In: CVPR (2022) 8
8. Cordts, M., Omran, M., Ramos, S., Rehfeld, T., Enzweiler, M., Benenson, R., Franke, U., Roth, S., Schiele, B.: The cityscapes dataset for semantic urban scene understanding. In: CVPR (2016) 8
9. Côté, G., Mannan, F., Thibault, S., Heide, F.: The differentiable lens: Compound lens search over glass surfaces and materials for object detection. In: CVPR. pp. 20803–20812 (2023) 7
10. Dai, J., Chen, L., Yang, X., Hu, Y., Gu, J., Xue, T.: Tolerance-aware deep optics. arXiv preprint arXiv:2502.04719 (2025), https://arxiv.org/abs/2502.04719 1, 7
11. Diamond, S., Sitzmann, V., Julca-Aguilar, F., Boyd, S., Wetzstein, G., Heide, F.: Dirty pixels: Towards end-to-end image processing and perception. ACM TOG **40**(3) (2021) 8
12. Geiger, A., Lenz, P., Urtasun, R.: Are we ready for autonomous driving? the KITTI vision benchmark suite. In: CVPR (2012) 8
13. Gharbi, M., Chaurasia, G., Paris, S., Durand, F.: Deep joint demosaicking and denoising. ACM TOG **35**(6) (2016) 8
14. Henz, B., Gastal, E.S.L., Oliveira, M.M.: Deep joint design of color filter arrays and demosaicing. Comput. Graph. Forum **37**(2), 389–399 (2018) 8
15. Jiang, Q., Shi, H., Gao, S., Zhang, J., Yang, K., Sun, L., Ni, H., Wang, K.: Computational imaging for machine perception: Transferring semantic segmentation beyond aberrations. arXiv preprint arXiv:2211.11257 (2024), https://arxiv.org/abs/2211.11257 1, 8
16. Liao, Y., Xie, J., Geiger, A.: KITTI-360: A novel dataset and benchmarks for urban scene understanding in 2D and 3D. IEEE TPAMI (2022) 1, 8
17. Long, J., Shelhamer, E., Darrell, T.: Fully convolutional networks for semantic segmentation. In: CVPR (2015) 8

18. Metzler, C.A., Ikoma, H., Peng, Y., Wetzstein, G.: Deep optics for single-shot high-dynamic-range imaging. In: CVPR. pp. 1375–1385 (2020) 7
19. Neuhold, G., Ollmann, T., Rota Bulò, S., Kontschieder, P.: The mapillary vistas dataset for semantic understanding of street scenes. In: ICCV (2017) 8
20. Ren, Z., Zhou, J., Zhang, W., Yan, J., Chen, B., Feng, H., Chen, S.: Successive optimization of optics and post-processing with differentiable coherent PSF operator and field information. arXiv preprint arXiv:2412.14603 (2024) 1, 7
21. Sakaridis, C., Dai, D., Van Gool, L.: Semantic foggy scene understanding with synthetic data. IJCV **126**(9) (2018) 8
22. Sakaridis, C., Dai, D., Van Gool, L.: Guided curriculum model adaptation and uncertainty-aware evaluation for semantic nighttime image segmentation. In: ICCV (2019) 8
23. Sakaridis, C., Dai, D., Van Gool, L.: ACDC: The adverse conditions dataset with correspondences for semantic driving scene understanding. In: ICCV (2021) 1, 8
24. Shah, I.A., Li, J., Delaney, E., Ward, E., Glavin, M., Jones, E., Deegan, B.: Learnable quantum efficiency filters for urban hyperspectral segmentation. arXiv preprint arXiv:2603.26528 (2026) 2, 6, 8, 13
25. Shah, I.A., Li, J., George, R., Brophy, T., Ward, E., Glavin, M., Jones, E., Deegan, B.: Hyperspectral sensors and autonomous driving: Technologies, limitations, and opportunities. arXiv preprint arXiv:2508.19905 (2025) 2, 6, 8, 13
26. Sitzmann, V., Diamond, S., Peng, Y., Dun, X., Boyd, S., Heidrich, W., Heide, F., Wetzstein, G.: End-to-end optimization of optics and image processing for achromatic extended depth of field and super-resolution imaging. ACM TOG **37**(4) (2018) 7
27. Strudel, R., Garcia, R., Laptev, I., Schmid, C.: Segmenter: Transformer for semantic segmentation. In: ICCV (2021) 8
28. Xie, E., Wang, W., Yu, Z., Anandkumar, A., Alvarez, J.M., Luo, P.: SegFormer: Simple and efficient design for semantic segmentation with transformers. In: NeurIPS (2021) 1, 8
29. Yang, X., Fu, Q., Heidrich, W.: Curriculum learning for ab initio deep learned refractive optics. Nature Communications **15**(1), 6572 (2024). https://doi.org/10.1038/s41467-024-50835-7 1, 3, 7
30. Yang, X., Fu, Q., Nie, Y., Heidrich, W.: Image quality is not all you want: Task-driven lens design for image classification. arXiv preprint arXiv:2305.17185 (2023) 1, 6, 8, 12
31. Yu, F., Chen, H., Wang, X., Xian, W., Chen, Y., Liu, F., Madhavan, V., Darrell, T.: BDD100K: A diverse driving dataset for heterogeneous multitask learning. In: CVPR (2020) 8
32. Zhao, H., Shi, J., Qi, X., Wang, X., Jia, J.: Pyramid scene parsing network. In: CVPR (2017) 8